\documentclass[journal]{IEEEtran}
\ifCLASSINFOpdf
\else
\fi
\usepackage{graphicx}
\usepackage{booktabs}
\usepackage{multirow}
\usepackage{color}

\begin{document}
\title{Twice Mixing: A Rank Learning based Quality Assessment Approach for Underwater Image Enhancement}

\author{Zhenqi Fu,
        Xueyang Fu,
        Yue Huang,
        and Xinghao Ding
\thanks{Zhenqi Fu, Yue Huang, and Xinghao Ding, are with the School of Informatics Xiamen University, Xiamen, 361005, China (e-mail: fuzhenqi@stu.xmu.edu.cn, yhuang2010@xmu.edu.cn, dxh@xmu.edu.cn).}
\thanks{Xueyang Fu is with School of Information Science and Technology, University of Science and Technology of China, Hefei, 230026, China (e-mail: xyfu@ustc.edu.cn).}}

\markboth{}%
{Shell \MakeLowercase{\textit{Fu et al.}}: Quality Assessment for Underwater Image Enhancement: A Rank Learning based Approach}

\maketitle

\begin{abstract}
To improve the quality of underwater images, various kinds of underwater image enhancement (UIE) operators have been proposed during the past few years. However, the lack of effective objective evaluation methods limits the further development of UIE techniques. In this paper, we propose a novel rank learning guided no-reference quality assessment method for UIE. Our approach, termed Twice Mixing, is motivated by the observation that a mid-quality image can be generated by mixing a high-quality image with its low-quality version. Typical mixup algorithms linearly interpolate a given pair of input data. However, the human visual system is non-uniformity and non-linear in processing images. Therefore, instead of directly training a deep neural network based on the mixed images and their absolute scores calculated by linear combinations, we propose to train a Siamese Network to learn their quality rankings. Twice Mixing is trained based on an elaborately formulated self-supervision mechanism. Specifically, before each iteration, we randomly generate two mixing ratios which will be employed for both generating virtual images and guiding the network training. In the test phase, a single branch of the network is extracted to predict the quality rankings of different UIE outputs. We conduct extensive experiments on both synthetic and real-world datasets. Experimental results demonstrate that our approach outperforms the previous methods significantly.

\end{abstract}


\IEEEpeerreviewmaketitle

\section{Introduction}
\IEEEPARstart{A}{s} an important carrier of oceanic information, underwater images play a critical role in ocean developments and explorations. For example, autonomous underwater vehicle and surveillance systems are usually equipped with an optical sensor for visual inspections and environmental sensing \cite{1}. Unfortunately, the captured images in underwater scenes are commonly degraded due to the wavelength-dependent light absorption and scattering. To achieve the requirement of high-quality underwater image generation, various underwater image enhancement (UIE) algorithms have been developed over the last few years \cite{2}.

Generally, the existing UIE algorithms can be organized into three categories: model-free, prior-based and data-driven methods. Model-free approaches, such as contrast limited adaptive histogram equalization (CLAHE) \cite{3}, Retinex \cite{4}, white balance \cite{5}, and Fusion \cite{6,7}, directly adjust pixel values without modeling the underwater degradation procedure. In contrast, prior-based methods restore degraded images based on elaborated physical imaging models and various prior knowledge. Dark Channel Prior (DCP) \cite{8} is one of the most adopted prior modes in UIE \cite{9,10,50}. Besides, another line of prior-based methods is to utilize the optical properties of underwater imaging. For example, Carlevaris-Bianco et al. \cite{11} used the difference of light attenuation to calculate the transmission map with the prior knowledge that red lights attenuate faster than the green and blue in the water environment. Recently, deep learning has shown remarkable success in various low-level and high-level vision tasks. Based on the powerful representations learned from a large quantity of annotated data, many researches on data-driven UIE algorithms have been presented \cite{12,13}.

\begin{figure*}
	\centering
	\includegraphics[scale=0.39]{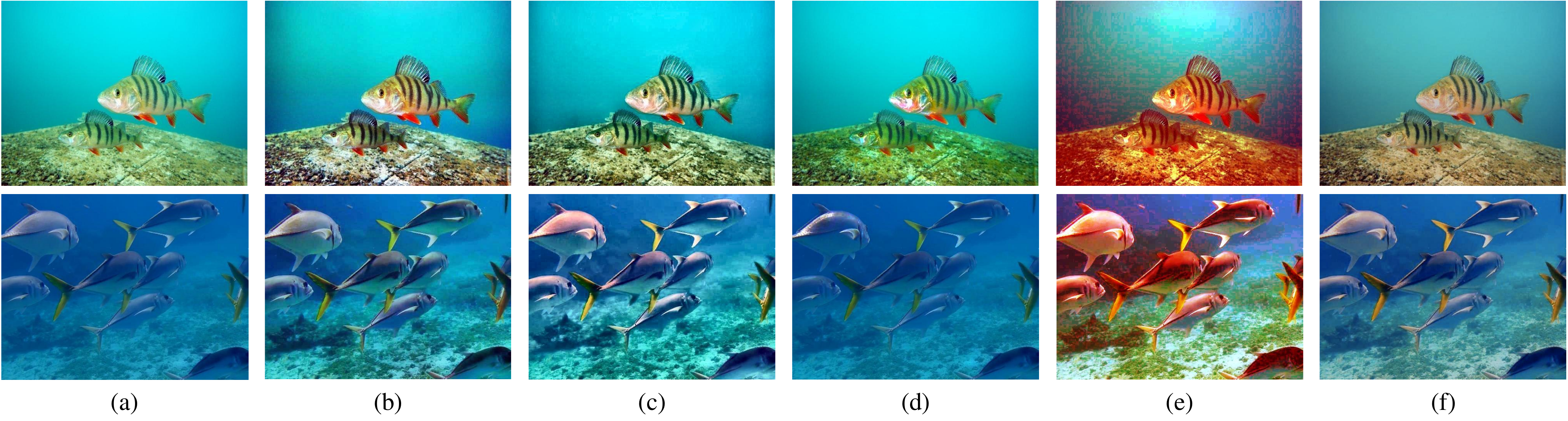}
	\caption{ Examples of enhanced underwater images. (a) Raw images, (b) CLAHE \cite{3}, (c) Fusion \cite{6}, (d) DCP \cite{8}, (e) Histogram \cite{15}, and (f) DuwieNet \cite{14}.}
	\label{FIG:1}
\end{figure*}

Although the existing UIE methods achieve impressive results, it is still unclear whether an underwater image with specific distortions can be successfully enhanced. As shown in Fig. 1, different UIE methods have their advantages and disadvantages in color correction and visibility improvement. Due to the lack of effective objective underwater image enhancement quality assessment (UIE-IQA) metrics, existing UIE methods rely on subjective comparisons to demonstrate the superiority of enhancement \cite{2,14}. However, the manual comparison takes a lot of time and effort. Even worse, subjective quality assessment is difficult to be integrated into online optimization systems to obtain real-time feedback on image quality. To ensure the enhanced images are perfect and satisfactory for real applications, objective UIE-IQA metrics should be performed, which have a direct application in optimizing the UIE operator based on the supervision of quality.

UIE-IQA belongs to blind/no-reference IQA (NR-IQA) \cite{47} because the reference image is unavailable in underwater scenes. However, directly using existing NR-IQA methods for UIE-IQA is inappropriate and difficult to achieve satisfactory results. This is because the quality degradation (e.g., color shift, contrast distortions, and artifacts) during the UIE is quite different from those in traditional NR-IQA metrics (e.g., noise, blur, and compression) \cite{16,17,18}. Previous works attempt to solve the UIE-IQA problem by empirically fusing different quality components \cite{42, 43}. Unfortunately, since the distortions of enhanced underwater images are too complex and affected by many factors, it is difficult to find a universal method by using only hand-crafted features with a shallow pooling module.

In this paper, we aim at designing learning-based approaches for UIE-IQA. Generally, learning-based quality predictors depend on amounts of annotated training data. Unfortunately, collecting quality annotations of enhanced underwater images for training a deep neural network is difficult. Because human observers can't give precise subjective judgments for such a large quantity of examples. To cope with this obstacle, we present a rank learning framework based on an elaborately designed self-supervision mechanism. The key idea is to randomly generate two mixing ratios before each iteration. The mixing ratios are utilized for both generating training data and supervise the training procedure. Concretely, before each iteration, we first mix the raw underwater image with its enhanced version twice following the random mixing ratios. Then, we adopt the mixed instances and their corresponding mixing ratios to train a Siamese Network \cite{19,20}. Our approach is inspired by the mixup algorithm \cite{51}, which linearly interpolates between two random input data and applies the mixed data with the corresponding soft label for training. The difference is that we simultaneously produce two mixed examples and train a Siamese Network to learn their rankings. This is because the perceptual quality of images non-linearly descends as the degradation increase. Therefore, we propose to apply the relative quality rather than the absolute score to train the network. Note that, the only subjective work for training our model is to collect raw underwater images, and their high-quality and low-quality enhanced versions. We do not use any subjective quality scores in the training phase. In summary, the main contributions of this paper are as follows:
\begin{itemize}
	\item [1)] 
	We present a rank learning framework for UIE-IQA based on an elaborately designed self-supervision mechanism \footnote{The dataset and code are available at: https://github.com/zhenqifu/Twice-Mixing.}. It is also the first time that using deep learning approaches to address the UIE-IQA problem. The core idea of our method is to randomly generate two mixing ratios, which are utilized for both generating input data and guiding the network training.
    \item [2)]
    We construct a dataset with over 2,200 raw underwater images and their high-quality and low-quality enhanced versions. Note that, our method is independent of subjective scores. The only subjective work is to collect high-quality and low-quality enhanced versions, which are easier to implement.
    \item [3)]	
    Extensive experiments on both synthetic and real-world UIE-IQA databases demonstrate that our method outperforms other methods significantly, and is more suitable for real applications. 
\end{itemize}

The rest of this paper is organized as follows. In section II, we briefly review the literature related to our work. In section III, we detail the proposed approach. We present the experimental results and the discussions in section IV. Finally, the conclusions are drawn in section V.

\section{Related Work}
In this section, we review the previous works related to this paper. We will first summarize the existing UIE methods, then we introduce the previous works of UIE-IQA.

\subsection{Underwater Image Enhancement}

Scattering and absorption effects caused by the water medium degrade the visual quality of underwater images and limit their applicability in vision systems. To improve the image quality, a lot of UIE algorithms have been developed. As mentioned above, those methods can be roughly classified into the following three categories. 

The first category is model-free methods, which focus on enhancing specific distortions via directly adjusting pixel values, without explicitly modeling the degradation procedure.  Traditional contrast limited adaptive histogram equalization (CLAHE) \cite{3}, histogram equalization (HE) \cite{53}, Retinex \cite{4} and white balance (WB) \cite{5} are several representative model-free algorithms. In literature \cite{21}, Iqbal et al. directly adjusted the dynamic range of pixels to improve the saturation and contrast. Ancuti et al. \cite{6} presented a fusion-based UIE method, in which a multi-scale fusion strategy is applied to fuse color corrected and contrast-enhanced images. An improvement version of \cite{6} is presented in \cite{7}, which adopts a white balancing technique and a novel fusing strategy to further promote the enhancement performance. Fu et al. \cite{4} proposed a retinex-based UIE approach, which contains three steps, i.e., color correction, layer decomposition, and post-processing. Ghani et al. \cite{22} devised a Rayleigh distribution guided UIE algorithm that can reduce the over/under-enhancement phenomena. Fu et al. \cite{23}, presented a two-step approach that addressed the absorption and scattering problems by color correction and contrast enhancement, respectively. Gao et al. \cite{24} enhanced underwater images based on the features of imaging environments and the adaptive mechanisms of the fish retina.

The second category is prior-based methods, which adopt prior knowledge and physical imaging models to improve the image quality. Dark Channel Prior (DCP) \cite{8} is one of the most used prior modes in this category methods. For example, Chiang et al. \cite{9} adopted DCP to remove haze and employed a wavelength-dependent compensation algorithm to correct colors. Drews et al. \cite{10} used a modified DCP algorithm to enhance underwater images, showing better transmission map estimation than the original DCP. Peng et al. \cite{25} presented a Generalized Dark Channel Prior (GDCP) method by integrating an adaptive color correction algorithm. Song et al. \cite{26} estimated the transmission map of the red channel by a new Underwater Dark Channel Prior (NUDCP). Apart from DCP, another line of prior-based algorithms is to apply the optical properties of underwater imaging. For instance, Galdran et al. \cite{27} recovered the colors associated with short wavelengths to enhance the image contrast. Zhao et al. \cite{28} computed the inherent optical properties of water medium from background colors. Li et al. \cite{15} utilized the histogram distribution prior and minimum information loss principle to enhance underwater images. Peng et al. \cite{29} first estimated the scene depth based on light absorption and image blurriness, then the depth information is employed for UIE. Berman et al. \cite{30} converted the issue of UIE to single image dehazing by predicting two additional global parameters. Wang et al. \cite{31} restored underwater images by combining the characteristics of light propagation and adaptive attenuation-curve prior.

The third category is data-driven based methods. Different from model-free and prior-based algorithms that apply hand-crafted features for UIE, data-driven approaches utilize the powerful modeling capabilities of deep learning to automatically extract representations and learn a nonlinear mapping from raw underwater images to the corresponding clean versions. Li et al. \cite{12} developed a deep learning based UIE method named WaterGAN. First, the authors simulated realistic underwater images in an unsupervised pipeline. Then, a two-stage network was trained end-to-end with the synthetic data. Li et al. \cite{13} proposed a weakly supervised UIE method using a cycle-consistent adversarial network \cite{32}, which relaxes the need for paired training data. Hou et al. \cite{33} jointly learned restoration information on transmission and image domains. Uplavikar et al. \cite{34} presented a novel UIE model based on adversarial learning to handle the diversity of water types. Jamadandi et al. \cite{35} exploited wavelet pooling and un-pooling to enhance degraded underwater images. Li et al. \cite{14} presented a fusion-based deep learning model for UIE using real-word images, the reference image is generated from twelve enhancement methods. Islam et al. \cite{36} presented a real-time UIE approach based on the conditional generative adversarial network. Fu et al. \cite{37} combined the merits of a traditional image enhancement technique and deep learning to improve the quality of underwater images. Guo et al. \cite{38} improved the quality of underwater images by a multi-scale dense generative adversarial network. Other relevant works of data-driven UIE methods can be found in \cite{39,40,41}. 

\subsection{Underwater Image Enhancement Quality Assessment}

Objective quality assessment for enhanced underwater images is a fundamentally important issue in UIE. However, it has not been deeply investigated. Researches on UIE-IQA do not keep pace with the rapid development of recent UIE methods. 

To objectively measure the quality of enhanced underwater images, Panetta et al. \cite{42} presented a linear combination based quality evaluation metric named UIQM, in which three individual quality measurements are devised to evaluate the colorfulness, sharpness, and contrast. UIQM is expressed as:
\begin{equation}
{\rm{UIQM}} = {c_1} \times {\rm{UICM}} + {c_2} \times {\rm{UISM}} + {c_3} \times {\rm{UIConM}}
\end{equation}
where UICM, UISM, and UIConM are the measurements of colorfulness, sharpness, and contrast, respectively. ${c_1}$, ${c_2}$ and ${c_3}$ are the weight factors dependent on the real applications. Yang et al. \cite{43} developed an UIE-IQA metric named UCIQE that calculates the standard deviation of chroma and the contrast of luminance in CIELab color space, and computes the average of saturation in HSV color space. The final quality score is obtained by a linear combination, which can be expressed as:
\begin{equation}
{\rm{UCIQE}} = {c_1} \times {\sigma _c} + {c_2} \times co{n_l} + {c_3} \times {\mu _s}
\end{equation}
where ${\sigma _c}$ is the standard deviation of chroma. ${co{n_l}}$ is the contrast of brightness.  ${\mu _s}$ is the average of saturation. ${c_1}$, ${c_2}$ and ${c_3}$ are the weight factors dependent on the real applications. Recently, Liu et al.\cite{2} constructed a large-scale real-world underwater image dataset to study the image quality, color casts, and higher-level detection/classification ability of enhanced underwater images. In literature \cite{14}, the authors built a UIE benchmark dataset named UIEBD. The dataset includes 890 real-world underwater images. To obtain the potential reference versions of each raw image, the authors first employed twelve UIE methods to generate enhanced versions. Then, the reference images of 890 original images were captured according to time-consuming and laborious pairwise comparisons. UIEBD provides a platform for designing data-driven UIE methods since it contains a lot of annotated data. Based on the dataset, the authors conducted a comprehensive study of the state-of-the-art algorithms qualitatively and quantitatively. Other relevant papers related to UIE-IQA can be found in \cite{44,45}.

\section{Proposed Method}

In this section, we introduce our approach to exploit rankings for UIE-IQA. We first lay out the framework of our approach and describe how we use a Siamese Network architecture to learn from rankings. Then we describe the dataset built for training the ranker.

\begin{figure*}
	\centering
	\includegraphics[scale=0.47]{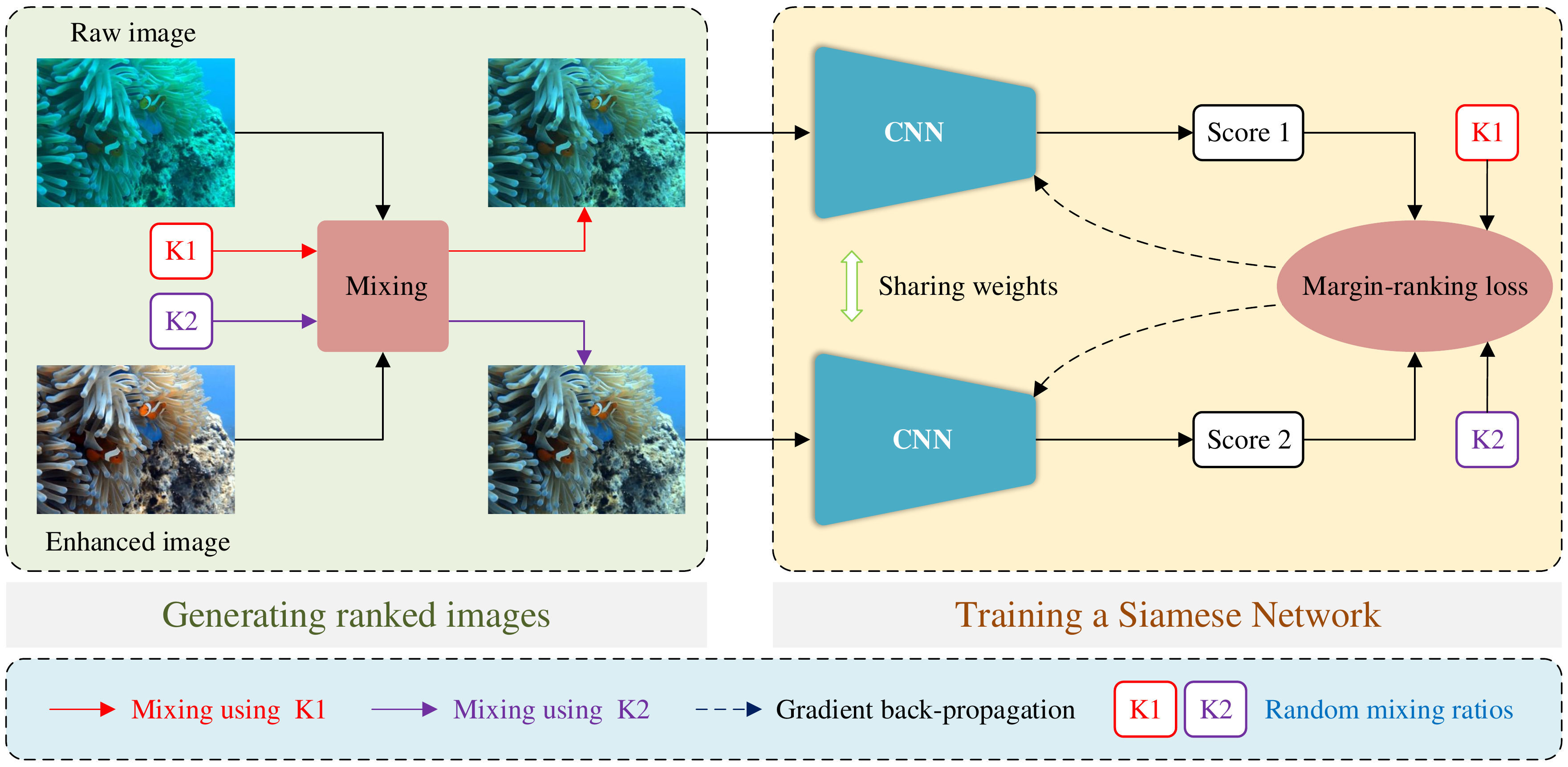}
	\caption{ \textbf {Overview of the proposed UIE-IQA method.} \textbf {In the training phase:} We randomly generate two mixing ratios before each iteration. Then pair-wise ranked images are implicitly synthesized and input into the network. The outputs of two branches are passed to the loss module, where we can compute the gradients based on two mixing ratios and apply back-propagation to update parameters of the whole network. \textbf {In the testing phase:} We extract a single branch from the network to predict the image quality.}
	\label{FIG:2}
\end{figure*}

\subsection{Generating Ranked Images}
To overcome the issue of lacking sufficient and effective training data, we formulate a simple yet reliable data augmentation approach that can automatically produce annotated examples. The core idea is to simultaneously generate pair-wise virtual images whose quality rankings are known in advance. Our method is inspired by the mixup algorithm \cite{51} which constructs virtual training examples by:
\begin{equation}
\left\{ \begin{array}{l}
\tilde x = \lambda {x_i} + \left( {1 - \lambda } \right){x_j}\\
\tilde y = \lambda {y_i} + \left( {1 - \lambda } \right){y_j}
\end{array} \right.
\end{equation}
where $x_i$ and $x_j$ are raw input vectors, $y_i$ and $y_j$ are labels (e.g. one-hot encodings for classification tasks), $\lambda$ denote the mixing ratio.
 
Nevertheless, Eq. 3 is invalid in UIE-IQA tasks. Because the human visual system is non-uniformity and non-linear in processing images. Linear interpolations of feature vectors might not lead to linear interpolations of the associated quality scores. Therefore, training a deep neural network to directly estimate the quality score is impracticable because we cannot obtain the ground-truth of each virtual instance. 

To address this problem, we develop a twice mixing strategy to simultaneously generate pair-wise virtual examples. Concretely, we first randomly produce two mixing ratios before each iteration. Then we perform the mixing module twice to produce two virtual training examples, as shown in Fig. 2. Mathematically, the mixed images can be formalized as: 

\begin{equation}
\left\{ \begin{array}{l}
{{\tilde x}_1} = {K_1}{x_i} + \left( {1 - {K_1}} \right){x_j}\\
{{\tilde x}_2} = {K_2}{x_i} + \left( {1 - {K_2}} \right){x_j}
\end{array} \right.
\end{equation}
where $x_i$ and $x_j$ are two input images, respectively. Without losing generality, we suppose the quality of $x_i$ is higher than $x_j$. $K_1$ and $K_2$ are mixing ratios that are randomly sampled from a uniform distribution before each iteration. ${K_1} \ne {K_2}$, and $\left| {{K_1} - {K_2}} \right| \ge 0.1$, to guarantee the visual difference between mixed instances. Although we are unable to obtain the absolute quality value of each virtual instance, we can capture their quality rankings in the light of mixing ratios. For instance, assigning a larger mixing ratio to the low-quality image result in lower quality of the generated virtual instance. Therefore, the rankings of pair-wise virtual examples can be calculated by:

\begin{equation}
\left\{ \begin{array}{l}
Q\left( {{{\tilde x}_1}} \right) < Q\left( {{{\tilde x}_2}} \right),{K_1} < {K_2}\\
Q\left( {{{\tilde x}_1}} \right) > Q\left( {{{\tilde x}_2}} \right),{K_1} > {K_2}
\end{array} \right.
\end{equation}
where $Q\left(  \cdot  \right)$ denotes the function of image quality. We show an example of virtual images in Fig. 3. As we can observe, the quality of mixed images gradually improves as the mixing ratio increases, which demonstrates the effectiveness of our approach.

\begin{figure*}
	\centering
	\includegraphics[scale=0.6]{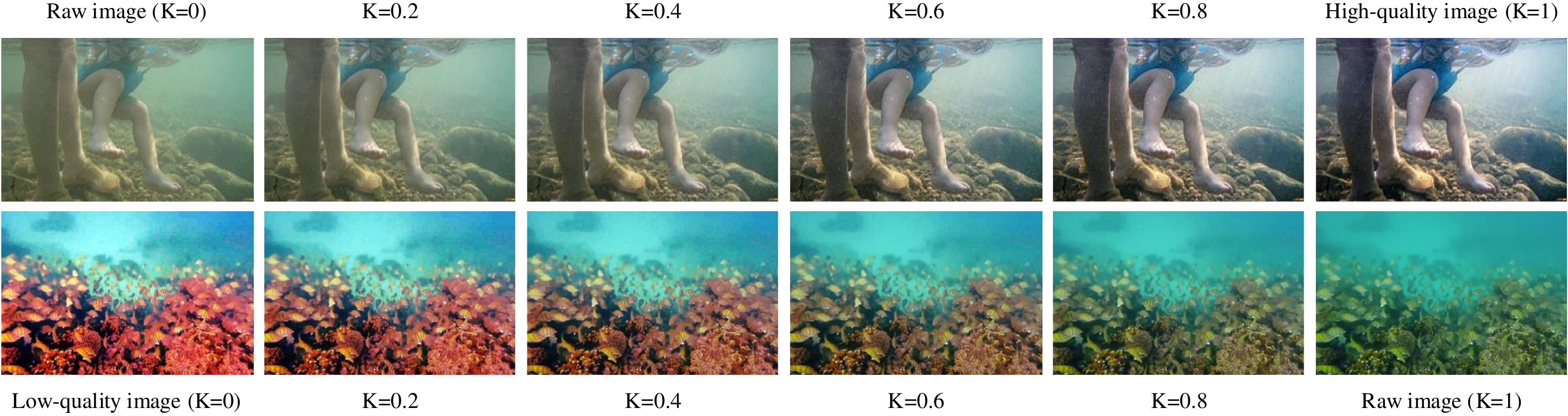}
	\caption{ Examples of generated virtual instances. The visual quality of mixed images gradually improves as the mixing ratio increases.}
	\label{FIG:3}
\end{figure*}

\subsection{Network Architecture}
Based on the proposed ranking data generating method, we introduce a network architecture to learn image quality from pair-wise ranked images. The whole framework of our approach is illustrated in Fig. 2. It contains two identical network branches and a loss module. In the training phase, the two branches are sharing weights. Pair-wise ranked images with associated labels (mixing ratios) are input to the network, yielding two quality representations. Here, we adopt a Global Average Pooling (GAP) layer after the feature extractor. Therefore the architecture can get rid of the limitation of input size. The quality representations are obtained after three fully-connected (FC) layers. The whole network is trained end-to-end effectively with a margin-ranking loss which will be described in the next subsection. We use VGG16 \cite{46} as the quality extractor that contains a series of convolutional, ReLU, and max-pooling layers. The output of the final layer is always a single scalar which is indicative of image quality. In the testing phase, we directly extract a single branch of the Siamese Network to predict the image quality. It is worth noting that our goal is to rank the enhanced underwater images instead of giving an absolute quality score, and we do not apply any subjective values in the training phase.

\subsection{Loss Function}

Different from most data-driven IQA methods that directly establish a nonlinear mapping from high-dimension feature space to low-dimension quality score space based on the powerful modeling capabilities of deep learning, we treat the quality prediction of enhanced underwater images as a sorting problem. Thus, the loss function employed in this paper is also different from traditional IQA tasks which address the quality evaluation as a regression issue. Specifically, we employ margin-ranking loss \cite{19,20} as the supervisor. Given two inputs ${\tilde{x}_1}$ and ${\tilde{x}_2}$, the output quality scores can be denoted by:
\begin{equation}
\left\{ \begin{array}{l}
{s_1} = f\left( {{\tilde{x}_1};\theta } \right)\\
{s_2} = f\left( {{\tilde{x}_2};\theta } \right)
\end{array} \right.
\end{equation}
where ${\theta}$ refers to the network parameters. Then the margin-ranking loss can be formulated as:
\begin{equation}
L\left( {{s_1},{s_2};\gamma } \right) = \max \left( {0,\left( {{s_1} - {s_2}} \right) * \gamma  + \varepsilon } \right)
\end{equation}
where ${s_1}$ and ${s_2}$ represent the estimated quality scores of ${\tilde{x}_1}$ and ${\tilde{x}_2}$,  respectively. The margin ${\varepsilon}$ is used to control the distance between ${s_1}$ and ${s_2}$. ${\gamma}$ is the ranking label of the pair-wise training images. ${\gamma}$ is computed by:

\begin{equation}
\left\{ {\begin{array}{*{20}{c}}
	{\gamma  = 1,}&{{Q\left( {{{\tilde x}_1}} \right)} < {Q\left( {{{\tilde x}_2}} \right)}}\\
	{\gamma  =  - 1,}&{Q\left( {{{\tilde x}_1}} \right) > Q\left( {{{\tilde x}_2}} \right)}
	\end{array}} \right.
\end{equation}

We replace $Q\left(  \cdot  \right)$ in Eq. 8 with Eq. 5, and obtain:
\begin{equation}
\left\{ {\begin{array}{*{20}{c}}
	{\gamma  = 1,}&{{K_1} < {K_2}}\\
	{\gamma  =  - 1,}&{{K_1} > {K_2}}
	\end{array}} \right.
\end{equation}

From Eq. 7 and Eq. 9, we can observe that the loss function utilized in this paper is only dependent on the two mixing ratios, which are randomly generated before each iteration. Therefore, the proposed UIE-IQA method is a self-supervision based model in essence. Finally, the ${n}$ pair-wise training images can be optimized by:

\begin{equation}
\begin{array}{l}
\hat \theta  = \mathop {\arg \min }\limits_\theta  \frac{1}{n}\sum\limits_{i = 1}^n {L\left( {s_1^{\left( i \right)},s_2^{\left( i \right)};{\gamma ^{\left( i \right)}}} \right)} \\
\;\;\; = \mathop {\arg \min }\limits_\theta  \frac{1}{n}\sum\limits_{i = 1}^n {L\left( {f\left( {x_1^{\left( i \right)};\theta } \right),f\left( {x_2^{\left( i \right)};\theta } \right);{\gamma ^{\left( i \right)}}} \right)} 
\end{array}
\end{equation}

\begin{figure*}
	\centering
	\includegraphics[scale=0.35]{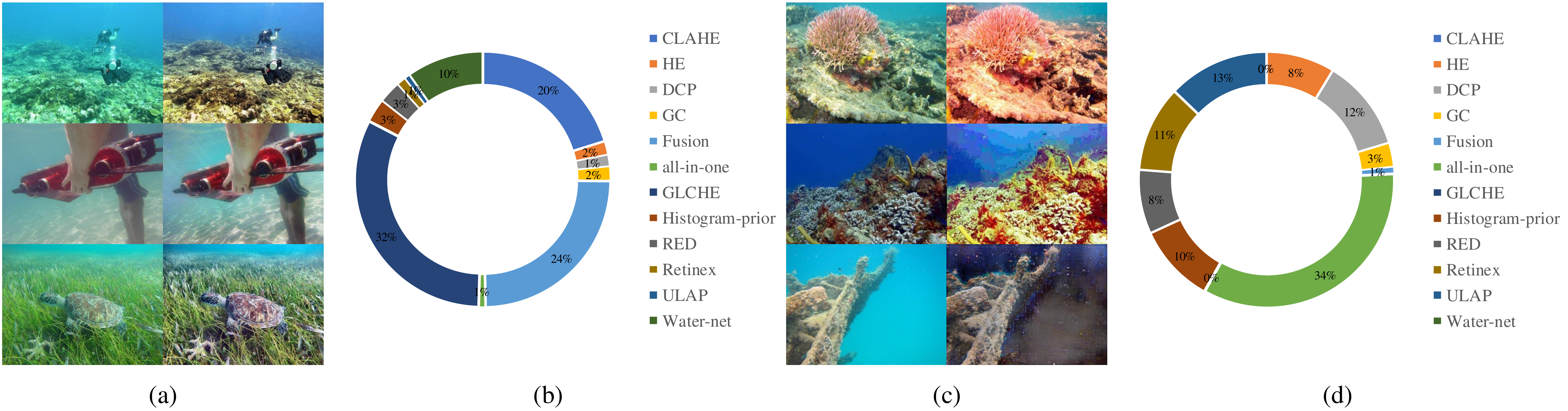}
	\caption{Examples of selected high-quality and low-quality enhanced underwater images. (a) High-quality underwater images. (b) The percentage of high-quality images from the results of different UIE algorithms. (c) Low-quality underwater images. (d) The percentage of low-quality images from the results of different UIE algorithms.}
	\label{FIG:4}
\end{figure*}

\subsection{Dataset}
Benefit from our elaborated self-supervision framework, the only subjective work for training our network is to collect high-quality (HQ) and low-quality (LQ) enhanced underwater images. Once those images are obtained, pair-wise ranked examples can be implicitly generated in the training procedure. To construct the dataset, we first collect 2,258 real-world underwater images, which have diverse scenes, different characteristics of quality degradation, and a broad range of image content. Then, we select 12 typical UIE methods, including five model-free methods (CLAHE \cite{3}, HE \cite{53}, Retinex \cite{4}, gamma correction (GC), and Fusion \cite{6}), four prior-based methods (DCP \cite{8}, ULAP \cite{52}, RED \cite{27}, and Histogram \cite{15}), and three data-driven methods (all-in-one \cite{34}, DuwieNet \cite{14}, and GLCHE \cite{37}). We employ these methods to generate enhanced underwater images. To ensure the quality of mixed instances span over a wide range of visual quality (from excellent to bad), and have diverse distortion types, the subjective selection is according to the following rules:
\begin{itemize}
	\item [i)] 
	The quality of HQ and LQ images should be higher and lower than corresponding original images, respectively.
	\item [ii)]
	The selected images should come from different UIE algorithms.
	\item [iii)] Choosing the best and the worst quality images as far as possible.  
\end{itemize}

\textbf{High-quality images:} It is easy to select the HQ images from different enhanced results since the quality of the enhanced version is better than the original one in most cases. Therefore, we obtain 2,258 HQ images. Fig. 3 (a) shows an example of HQ images in our dataset. The percentage of HQ images from the results of different UIE algorithms is presented in Fig. 3 (b). 

\textbf{Low-quality images:} However, LQ images are not always existing. After subjectively selecting, we obtain 1,815 LQ images. Examples of collected LQ images are shown in Fig. 3 (c). We can observe that the LQ images have significant over-enhancement effects (e.g., serious reddish color shift, excessive contrast, and over-saturation). Their visual quality is lower than the original images. We illustrate the percentage of LQ images from the results of different UIE methods in Fig. 3 (d).

\section{Experimental Results}

We test the model performance on the synthetic dataset as well as the real-world dataset. Next, we will first give our experimental settings, then we conduct several experiments to show the excellent improvements of our methods. 

\subsection{Experimental Settings}

\textbf {Parameters setting.} For our Siamese Network, we use VGG16  \cite{46} as the quality predictor, where the kernel size of all convolutional layers are 3 × 3, and after each convolutional layer, ReLU operation is used as the nonlinear mapping.  We set the final FC layer of VGG16 as 1. The parameter ${\varepsilon}$ is set as 0.5 empirically. We use the Pytorch framework to train our network utilizing the Adam solver with an initial learning rate of 1e-6. The mini-batch size is empirically set as 1.

\textbf {Training and testing datasets.}
We train the network on our newly constructed dataset. We use the first 2,000 original underwater images and their corresponding HQ and LQ images for training, and the rest for testing. Note that, in the testing phase, the mixing ratio $K$ is fixed. We set $K = 0,0.2,0.4,0.6,0.8$, to build the synthetic testing dataset. We also evaluate the model performance on PKU dataset \cite{44}, which contains 100 raw underwater water. For each source image, five UIE algorithms are applied to generate enhanced images. Subjective rankings of enhanced images are labeled by 30 volunteers.

\textbf {Compared methods.} Three state-of-the-art metrics including one traditional IQA algorithm (NIQE \cite{47}) and two UIE-IQA methods (UIQM \cite{42} and UCIQE \cite{43}) are employed for performance comparisons. We record the results of all competitors by conducting the same experiments using the original implementations provided by the authors. For UIQM and UCIQE, a larger value indicates better image quality, while a smaller value means better image quality for NIQE. 

\textbf {Performance Criteria.} We use Kendall Rank Correlation Coefficient (KRCC) \cite{48} and Spearman Rank Correlation Coefficient (SRCC) \cite{47} to assess the model’s performance. Formally, KRCC is defined as:
\begin{equation}
{\rm{KRCC}} = \frac{{{n_c} - {n_d}}}{{0.5n\left( {n - 1} \right)}}
\end{equation}
where ${n}$ is the ranking length. ${n_c}$ and ${n_d}$ are the number of concordant and discordant pairs, respectively. SRCC is defined as:

\begin{equation}
{\rm{SRCC}} = 1 - \frac{{6\sum {d_i^2} }}{{n\left( {{n^2} - 1} \right)}}
\end{equation}
where ${n}$ is the ranking length. $d_i$ denotes the difference in rankings of the element $i$. A better objective UIE-IQA measure is expected to get higher SRCC and KRCC.

\begin{table}[!t]
	\caption{Performance of different methods on the synthetic testing database. The best results are in bold.}\label{tbl1}
	\renewcommand\tabcolsep{5pt}
	\centering
	\begin{tabular}{cccccc}
		\toprule
		Dataset & Criteria & UIQM & UCIQE & NIQE & Twice Mixing \\	
		\midrule
		\multirow{4}*{Synthetic} & Mean KRCC & 0.1025 & 0.0397 & -0.0238 & \textbf{0.6718} \\
		~ & Std KRCC & 0.9447 & 0.9238 & 0.6664 & \textbf{0.5705}\\
		~ & Mean SRCC & 0.1027 & -0.0215 & -0.0822 & \textbf{0.6872}\\
		~ & Std SRCC & 0.9530 & 0.9369 & 0.7293 & \textbf{0.5928}\\								
		\bottomrule
	\end{tabular}
\end{table}

\begin{figure*}
	\centering
	\includegraphics[scale=0.32]{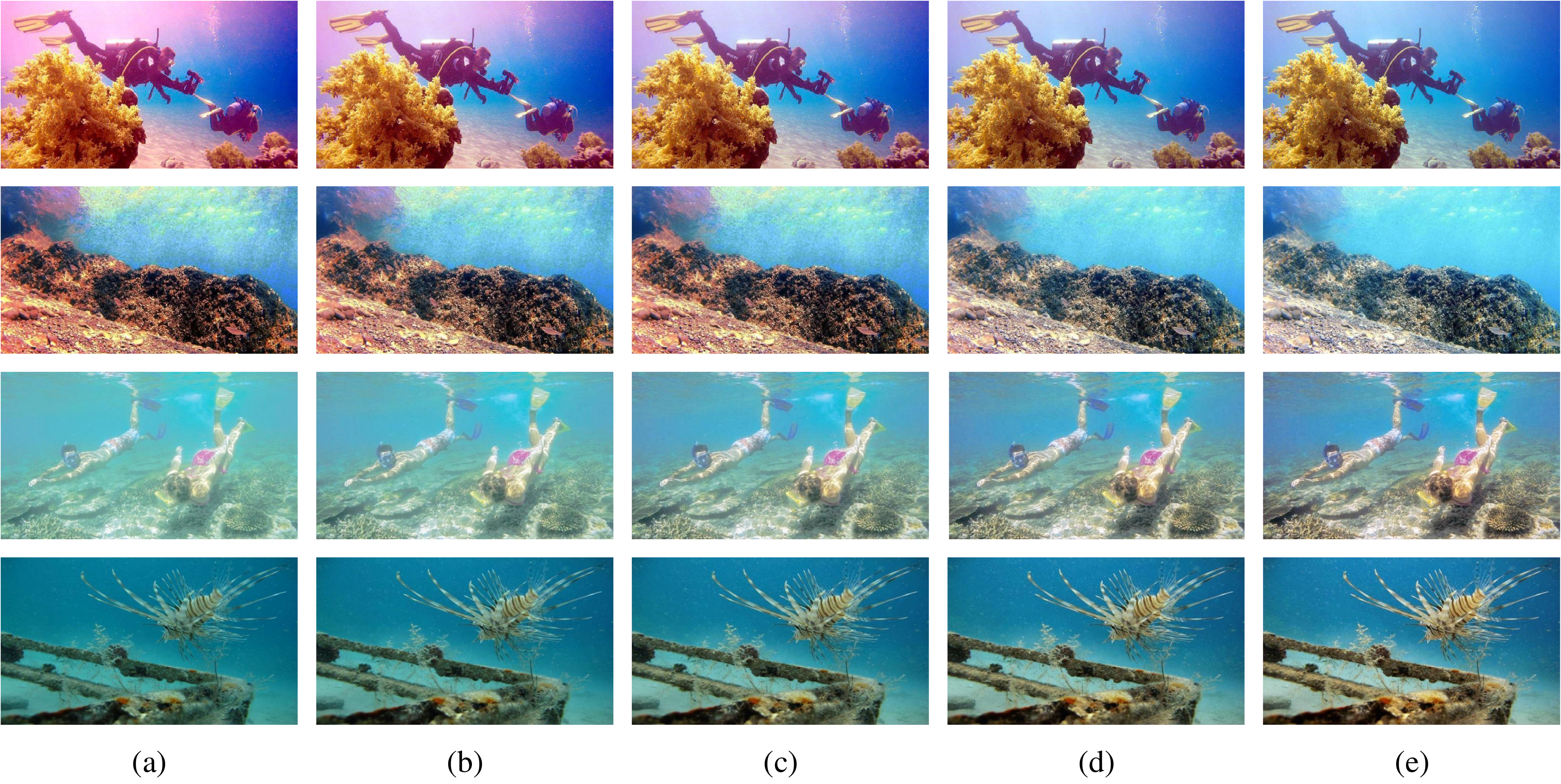}
	\caption{Underwater images in the synthetic testing dataset. The quality rankings of (a)-(e) gradually improve. Corresponding objective evaluation results are listed in Tab. II.}
	\label{FIG:5}
\end{figure*}

\begin{table*}[!t]
	\caption{Evaluation results of underwater images in Fig 5. Here, GT denotes the ground-truth ranking.}\label{tbl2}
	\renewcommand\tabcolsep{4pt}
	\centering	
	\begin{tabular}{ccccccccccccc}
		\toprule
		\multirow{2}{*}{Metric}  & \multicolumn{6}{c}{First line} & \multicolumn{6}{c}{Second line}\\ \cmidrule[0.05em](lr){2-7} \cmidrule[0.05em](l){8-13}
		&       (a) & (b) & (c) & (d) & (e)  & KRCC    & (a) & (b) & (c) & (d) & (e) & KRCC\\ 
		\midrule
		UIQM & 5.5322 & 5.1214 & 4.6711 & 4.1289 & 3.6437 & -1 & 4.0162 & 3.7113 & 3.3451 & 2.8354 & 2.2806 & -1\\
		UCIQE & 33.206 & 33.293 & 33.465 & 33.424 & 33.219 & 0.2 & 33.470 & 32.592 & 32.566 & 32.979 & 33.404 & 0\\
		NIQE & 2.0907 & 2.1450 & 2.1061 & 2.1315 & 2.1015 & 0 & 3.2759 & 3.2685 & 3.2336 & 3.0810 & 3.0405 & 1\\
		Twice Mixing & 0.8917 & 1.0074 & 1.1357 & 1.2870 & 1.3255 & 1 & 0.5726 & 0.6899 & 0.8295 & 0.9459 & 0.9766 & 1 \\
		GT & 1 & 2 & 3 & 4 & 5 & 1 & 1 & 2 & 3 & 4 & 5 &	1\\
		
		\bottomrule
		\toprule
		
		\multirow{2}{*}{Metric}  & \multicolumn{6}{c}{Third line} & \multicolumn{6}{c}{Fourth line}\\ \cmidrule[0.05em](lr){2-7} \cmidrule[0.05em](l){8-13}
		&       (a) & (b) & (c) & (d) & (e) & KRCC   & (a) & (b) & (c) & (d) & (e) & KRCC\\ 
		\midrule
		UIQM & 1.6375 & 2.0534 & 2.5076 & 3.0093 & 3.4110 & 1 & 0.2399 & 0.8361 & 1.2167 & 1.7660 & 2.0508 & 1\\
		UCIQE & 22.241 & 24.303 & 26.326 & 30.604 & 32.477 & 1 & 28.511 & 29.170 & 29.856 & 30.663 & 31.726 & 1\\
		NIQE & 3.0878 & 3.0809 & 3.0626 & 3.1264 & 3.1818 & -0.4 & 3.7144 & 3.8490 & 3.8205 & 3.2228 & 3.3172 & 0.4\\
		Twice Mixing & 0.9771 & 1.1423 & 1.2865 & 1.3871 & 1.4790 & 1 & 0.4713 & 0.5198 & 0.6207 & 0.7143 & 0.7568 & 1 \\
		GT & 1 & 2 & 3 & 4 & 5 & 1 & 1 & 2 & 3 & 4 & 5 &1	\\		
		\bottomrule
		
	\end{tabular}
\end{table*}

\subsection{ Performance Comparisons on Synthetic Data}

In this subsection, we verify the model performance on our newly built synthetic testing dataset, in which the ground-truth is automatically labeled according to the mixing ratio. Tab. I summarizes the comparison results of mean and standard deviation (std) KRCC and SRCC values. From the table, we can observe that the proposed method is significantly better than the other three metrics. Traditional NR-IQA metric NIQE shows the worst performance because NIQE is designed for terrestrial images and it does not take the specific distortions of UIE into account. Compared with two UIE-IQA methods, our approach utilizes the powerful modeling capabilities of deep learning to learn representative features from pair-wise samples, without manually designing specific low-level quality relevant features. Therefore, the proposed approach achieves better performance. Further, we provide an example to show the capability of different UIE-IQA metrics in predicting quality values. 

We test four groups of enhanced underwater images as shown in Fig. 5. The first and second groups have over-enhancement appearances, while examples in the third and fourth lines are under-enhancement. We report the evaluation results in Tab. II. From Fig. 5 and Tab. II, we can make the following observations:

\begin{itemize}
	\item [i)] 
	UIQM, UCIQE cannot accurately estimate the quality of over-enhanced images. Their results are even worse than the traditional NR-IQA method NIQE in this case. Especially for UIQM, the reddish results consistently exhibit the highest quality scores. We consider the reasons may lie in that UIQM and UCIQE are designed based on hand-crafted low-level features with a simple linear pooling, which limits their performance and generalizations. 

	\item [ii)] 	
	Since the quality degradations between enhanced underwater images and terrestrial instances are quite different. Directly applying traditional NR-IQA metrics cannot achieve satisfactory evaluation results. As reported in Tab. II, NIQE fails to predict the quality of under-enhanced images, while the other three UIE-IQA metrics can accurately estimate. 
		
	\item [iii)]	
	Compared with three competitors, our results are highly consistent with subjective rankings. This is benefits from our dataset that contains both high-quality and low-quality enhanced underwater images, and the elaborate formulated twice mixing strategy. Instead of manually designing features, Twice Mixing adopts a Siamese Network to learn the pair-wise comparison with a margin-ranking loss. As a result, the proposed model can automatically dig discriminative features that are more relevant to the image quality.  	
	
\end{itemize}

\begin{table}[!t]
	\caption{Performance of different methods on the PKU database. The best results are in bold.}\label{tbl3}
	\renewcommand\tabcolsep{5pt}
	\centering
	\begin{tabular}{cccccc}
		\toprule
		Dataset & Criteria & UIQM & UCIQE & NIQE & Twice Mixing \\
		\midrule	
		\multirow{4}*{PKU} & Mean KRCC & -0.5240 & 0.3800 & -0.0920 & \textbf{0.5920} \\
		~ & Std KRCC & 0.4356 & \textbf{0.3222} & 0.4675 & 0.3662 \\
		~ & Mean SRCC & -0.6060 & 0.4620 & -0.0840 & \textbf{0.6707} \\
		~ & Std SRCC & 0.4799 & \textbf{0.3757} & 0.5583 & 0.3971 \\
		\bottomrule
	\end{tabular}
\end{table}

\subsection{Performance Comparisons on PKU Dataset}

\begin{table*}[!t]
	\caption{Evaluation results of underwater images in Fig. 6. Here, GT denotes the ground-truth ranking.}\label{tbl4}
	\renewcommand\tabcolsep{4pt}
	\centering	
	\begin{tabular}{ccccccccccccc}
		\toprule
		\multirow{2}{*}{Metric}  & \multicolumn{6}{c}{First line} & \multicolumn{6}{c}{Second line}\\ \cmidrule[0.05em](lr){2-7} \cmidrule[0.05em](l){8-13}
		&       (a) & (b) & (c) & (d) & (e)  & KRCC   & (a) & (b) & (c) & (d) & (e) & KRCC \\ 
		\midrule
		UIQM & 2.2369 & -0.2278 & -2.5231 & 3.3535 & 3.8102 & -0.4 & 0.7315 & 0.4252 & 1.2921 & 2.1434 & 3.5726 & -0.8\\
		UCIQE & 31.932 & 29.084 & 16.375 & 28.355 & 24.286 & 0.6 & 36.317 & 33.492 & 37.694 & 31.957 & 33.198 & 0.4\\
		NIQE & 4.5148 & 6.2774 & 7.7976 & 5.2964 & 5.5633 & 0.2 & 3.3247 & 2.5724 & 4.0111 & 2.6841 & 2.7832 & 0 \\
		Twice Mixing & 0.5834 & 0.3471 & 0.2425 & -0.2008 & -0.4575 & 1 & 1.8558 & 1.7431 & 1.4316 & 1.3137 & 0.6898 & 1 \\
		GT & 5 & 4 & 3 & 2 & 1 &  1 & 5 & 4 & 3 & 2 & 1 & 1	\\
		
		\bottomrule
		\toprule
		
		\multirow{2}{*}{Metric}  & \multicolumn{6}{c}{Third line} & \multicolumn{6}{c}{Fourth line}\\ \cmidrule[0.05em](lr){2-7} \cmidrule[0.05em](l){8-13}
		&       (a) & (b) & (c) & (d) & (e)  & KRCC   & (a) & (b) & (c) & (d) & (e) & KRCC\\ 
		\midrule
		UIQM & 4.2602 & 4.1195 & 3.6208 & 0.0081 & 0.7238 & 0.8 & 1.2348 & 1.3499 & 2.7651 & 4.9258 & 4.0820 & -0.8\\
		UCIQE & 34.218 & 30.840 & 22.756 & 31.296 & 21.267 & 0.6 & 31.647 & 32.223 & 29.765 & 37.490 & 28.655 & 0.2\\
		NIQE & 3.9804 & 4.2074 & 4.2075 & 3.6032 & 4.1573 & 0 & 2.5766 & 3.0362 & 2.9796 & 3.2380 & 3.0023 & 0.4\\
		Twice Mixing & 1.9827 & 1.3385 & 0.7734 & 1.1111 & 0.6405 & 0.8 & 3.1275 & 3.7372 & 2.9139 & 2.3859 & 2.0343 & 0.8\\
		GT & 5 & 4 & 3 & 2 & 1 & 1 & 5 & 4 & 3 & 2 & 1 & 1	\\		
		\bottomrule
		
	\end{tabular}
\end{table*}

\begin{figure*}
	\centering
	\includegraphics[scale=0.39]{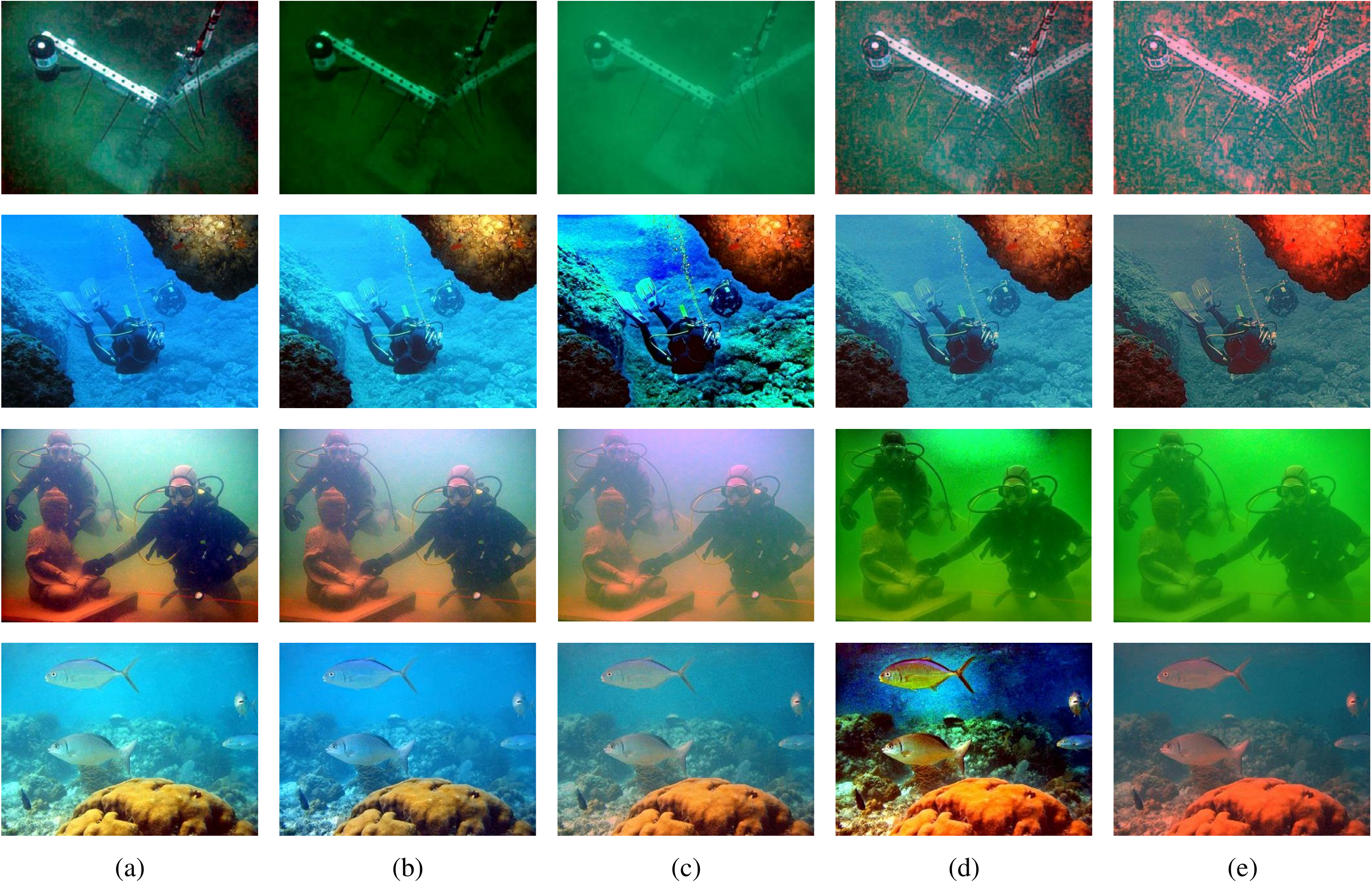}
	\caption{Underwater images in the PKU dataset. The quality rankings of (a)-(e) gradually deteriorate. Corresponding objective evaluation results are listed in Tab. IV.}
	\label{FIG:6}
\end{figure*}

The Holy Grail we pursue is to design a generic UIE-IQA metric that can predict enhanced underwater image quality robustly and accurately. Therefore, we further conduct experiments on the PKU dataset to show the model performance for real UIE outputs. PKU dataset is constructed in \cite{44}. It contains 100 original underwater images and 500 enhanced versions generated by five UIE algorithms. Subjective rankings are obtained via a well-designed user study. Tab. III gives the results of the comparisons. As shown in the table, our method achieves the best performance compared with UIQM, UCIQE, and NIQE. We note that UIQM acquires a high negative correlation with subjective rankings. This is because UIQM favors the outputs with over-enhancement effects. As expected, NIQE obtains poor results since it is designed for terrestrial images with specific distortions. 

To intuitively show the performance of each method, Tab. IV gives an example of quality estimation for enhanced underwater images presented in Fig. 6. We can make the following observations from Fig. 6 and Tab. IV:

\begin{itemize}
	\item [i)] 
	UIQM and UCIQE have their preferences for UIE-IQA tasks. Their results are not always subjectively correct. For example, UIE algorithms tend to produce excessive redness due to over-enhancement, resulting in extremely high UIQM scores. Unfortunately, these results are visually unfriendly according to human visual perceptions. For UCIQE, it also tends to produce high quality values for reddish results. Meanwhile, UCIQE favors the outputs with high contrasts, such as the enhanced instances of (d) in the third and fourth lines, and (c) in the second line. 
	
	\item [ii)]	
	Similar to the evaluation results on our synthetic testing dataset, NIQE cannot handle the UIE-IQA task well. As illustrated in Fig 6, the degradations of enhanced underwater images are mainly caused by the color cast, contrast distortions, naturalness, and so on, which is significantly different from the distortions in the traditional IQA community. Therefore, directly adopting traditional IQA metrics may fail to capture the desired results.
	
	\item [iii)]	
	UIQM, UCIQE, and NIQE use hand-crafted low-level features for quality evaluation, their performance is highly dependent on the designer's experience and the effectiveness of quality pooling strategies. To be more specific, UIQM and UCIQE should carefully balance each quality component that is manually extracted. Benefited from the powerful modeling capabilities of deep neural networks and the elaborately designed self-supervision mechanism, the proposed method can automatically and effectively learn the intricate relationship between enhanced underwater images and their subjective quality rankings from massive training data. As a result, the proposed method achieves better evaluation accuracy and generalizations compared with UIQM, UCIQE, and NIQE. 
	
	\item [iv)]
	Although Twice Mixing achieves state-of-the-art performance, there are still false estimates (e.g., third and fourth lines). We consider that UIE-IQA is a challenging task, which needs to comprehensively assess diverse distortions. Therefore, the development of an appropriate UIE-IQA metric is still an open issue in this field, and there is still much room for the improvement of UIE-IQA.

\end{itemize}

\section{Conclusion}

In this paper, we propose a rank learning framework for UIE-IQA based on an elaborately formulated self-supervision mechanism. The core idea is to randomly generate two mixing ratios, which are utilized for both generating training examples and corresponding rankings. Unlike typical mixup algorithms that calculate the annotations of virtual instances via a linear combination, by considering that the human visual system is non-uniformity and non-linear in processing images, we propose to compute quality rankings of two virtual instances according to the random mixing ratios. Therefore, we train a Siamese Network to learn the pair-wise comparison with a margin-ranking loss. To train our network, we construct a dataset with over 2,200 raw underwater images and their high-quality and low-quality enhanced versions. The performance of our metric is extensively verified by several elaborately designed experiments, on both synthetic and real-world UIE-IQA datasets. Experimental results show that the proposed method obtains superior performance compared to existing UIE-IQA techniques. The prediction results are in line with subjective judgments. In future works, we will focus on digging prior knowledge for quality representation and concentrating on accurate quality pooling. Also, we plan to design specific UIE algorithms under the guidance of UIE-IQA methods.

\bibliographystyle{IEEEtran}
\bibliography{IEEEabrv,cas-refs}

\end{document}